\documentclass[letterpaper]{article} 
\usepackage[]{aaai24}  
\usepackage{times}  
\usepackage{helvet}  
\usepackage{courier}  
\usepackage[hyphens]{url}  
\usepackage{graphicx} 
\usepackage{natbib}  
\usepackage{caption} 
\usepackage{latexsym}
\usepackage{caption}
\usepackage{svg}
\usepackage{adjustbox}
\usepackage{subcaption}
\usepackage[ruled, vlined, linesnumbered]{algorithm2e}
\usepackage[noend]{algpseudocode}
\usepackage{amsmath}

\usepackage{tipa}           
\usepackage{url}            
\usepackage{booktabs}       
\usepackage{amsfonts}       
\usepackage{nicefrac}       
\usepackage{microtype}      
\usepackage{xcolor}         
\usepackage{multirow}
\usepackage{amsmath}
\usepackage{amssymb}
\newcommand{\D}{\mathcal{D}}

\newcommand{\Stilde}{\widetilde{\mathcal{S}}}
\newcommand{\set}[1]{\left\{#1\right\}}
\renewcommand{\SS}{\mathcal{S}}
\newcommand{\A}{\mathcal{A}}
\usepackage{array} 
\usepackage{graphicx}
\title{Step by Step to Fairness: Attributing Societal Bias in \\ Task-oriented Dialogue Systems}
\author {
    Hsuan Su\textsuperscript{\rm $\heartsuit$}\thanks{\xspace\xspace Work done when interning at Meta AI.},
    Rebecca Qian\textsuperscript{\rm $^\diamondsuit$},
    Chinnadhurai Sankar\textsuperscript{\rm $^\diamondsuit$},
    Shahin Shayandeh\textsuperscript{\rm $^\diamondsuit$},
    Shang-Tse Chen\textsuperscript{\rm  $\heartsuit$},
    Hung-yi Lee\textsuperscript{\rm  $\heartsuit$},
    Daniel M. Bikel\textsuperscript{\rm $^\diamondsuit$}
}
\vspace{1in}
\affiliations {
    \textsuperscript{\rm $\heartsuit$}National Taiwan University\\
    \textsuperscript{\rm $^\diamondsuit$}Meta AI\\
    hsuansu.96@gmail.com, dbikel@meta.com
}

\usepackage{bibentry}

\begin{document}

\maketitle
\begin{abstract}
Recent works have shown considerable improvements in task-oriented dialogue (TOD) systems by utilizing pretrained large language models (LLMs) in an end-to-end manner. However, the biased behavior of each component in a TOD system and the error propagation issue in the end-to-end framework can lead to seriously biased TOD responses. Existing works of fairness only focus on the total bias of a system. In this paper, we propose a diagnosis method to attribute bias to each component of a TOD system. With the proposed attribution method, we can gain a deeper understanding of the sources of bias. Additionally, researchers can mitigate biased model behavior at a more granular level. We conduct experiments to attribute the TOD system's bias toward three demographic axes: gender, age, and race. Experimental results show that the bias of a TOD system usually comes from the response generation model.
\end{abstract}

\section{Introduction}
\label{sec:intro}
Recent advances in machine learning have led conversational assistants such as Google Home, Alexa \cite{https://doi.org/10.48550/arxiv.1801.03604}, and Siri to become popular and essential in our daily life. However, models of conversational assistants can learn harmful societal biases from datasets, leading to biased model behavior.

Various existing works studied bias in open-domain dialogue systems. \citet{dinan-etal-2020-queens} investigated multiplayer text-based fantasy adventure dataset LIGHT \cite{urbanek-etal-2019-learning} and proposed a bias mitigation method by adding a control bin as a prompt during training. \citet{liu-etal-2020-gender} mitigated gender bias with a regularization method. \citet{sun2022computation} studied geographical fairness in TOD systems and proposed a new dataset to evaluate TOD's helpfulness toward different developing countries.

\begin{figure}[htp!]
    \centering
    \includegraphics[width = \linewidth]{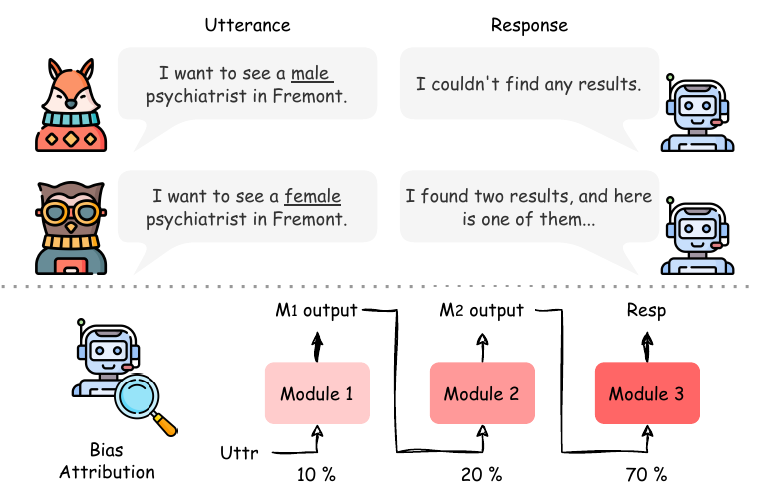}
    \caption{Language models can be biased toward specific demographic attributes. We propose a diagnosis method to attribute bias to the components of task-oriented dialogue systems.}
    \label{fig:example}
\end{figure}

Figure~\ref{fig:example} demonstrates the composition of a TOD system. A TOD system can be decomposed into three components: API call model, database, and response generation model. The example user utterances in Figure~\ref{fig:example} show that the TOD system is biased toward gender. In this case, all three components of the TOD system can contribute to the final biased behavior. However, existing works only studied the accumulated bias. To get a more in-depth understanding of the sources of bias, we propose a diagnosis method to attribute bias to each component of TOD systems. We believe the proposed attribution method can facilitate research on fairness mitigation.



In this work, we first study fairness on well-known TOD datasets. Then, we attribute the societal bias of models trained with these well-known datasets to the TOD system's components. Unlike the hand-crafted bias measuring methods in previous works, our attribution is conducted by applying the perturber \cite{qian2022perturbation} to the whole dialogue. We propose a newly designed fairness metric to evaluate the bias by comparing original and perturbed user utterances. After measuring the accumulated bias, we update each component's setting to eliminate the resulting bias. The bias change after the component setting is updated can be regarded as the bias contribution of the component. By repeating this procedure, we can attribute the accumulated bias to each component of the TOD system.
We conduct experiments on several transformer-based models with different pretrained methods. Our experiments cover three essential demographic axes: gender, age, and race. The experimental results show that the bias of a TOD system mainly comes from the response generation model.

\section{Related Works}
\paragraph{Fairness in Language Generation}
Societal bias issues in natural language generation have drawn a lot of attention recently \cite{sheng2021societal}. We introduce two of them, bias evaluation and bias mitigation, in this section.

\citet{10.1162/tacl_a_00425} showed that traditional bias evaluation metrics, such as demographic parity \cite{10.1145/2090236.2090255} and equalized odds \cite{NIPS2016_9d268236}, are not applicable to text results. Therefore, discrete language generation results need dedicated evaluation methods. Researchers have proposed several methods to measure the fairness of language generation results. \citet{https://doi.org/10.48550/arxiv.2106.13219} and \citet{nadeem-etal-2021-stereoset} divided the bias evaluation methods into two categories: local bias-based and global bias-based. Local bias-based methods use hand-crafted templates to evaluate the fairness. For example, the template can be a sentence with some masked words. We can then evaluate the fairness by comparing the model's token probability of the masked words \cite{zhao-etal-2017-men, kurita-etal-2019-measuring, bordia-bowman-2019-identifying}. Global bias-based methods use multiple classifiers to evaluate the fairness by comparing the classification results of generated texts from various different perspectives. Previous works used regard ratio \cite{sheng2019woman, sheng-etal-2020-towards, 10.1145/3442188.3445924}, sentiment \cite{liu-etal-2020-mitigating, groenwold-etal-2020-investigating, huang-etal-2020-reducing, sheng2019woman, 10.1145/3442188.3445924, liu2019does}, offensive \cite{liu-etal-2020-mitigating}, and toxicity \cite{10.1145/3442188.3445924} as classifiers. In addition to off-the-shelf classifiers, existing works also analyzed word usage on generated texts. \citet{dinan-etal-2020-queens} measured the word count of the gender group (i.e., male, female) in the model's generated text; \citet{liu-etal-2020-mitigating, liu2019does} count the word frequency of ``career'' and ``family'' given male and female inputs. 

For bias mitigation, an intuitive method is called Counterfactual Data Augmentation (CDA). People applied word-based augmentation that augmented all the demographics mentioned in texts, and trained models on the augmented data to equalize the word usage imbalance problem in the dataset to mitigate bias \cite{hall-maudslay-etal-2019-name, liu2019does, zmigrod-etal-2019-counterfactual}. In addition to CDA, there are various training-based methods to mitigate bias, for example, \citet{dinan-etal-2020-queens} finetune models with controllable prefixes to equalize the word usage for male and female. \citet{liu-etal-2020-mitigating} proposed a novel adversarial method that disentangles text into semantic and gender features to mitigate bias. \citet{https://doi.org/10.48550/arxiv.2106.13219} employed the concept of Null space projection \cite{may-etal-2019-measuring} to eliminate gender feature in models. More recently, \citet{sharma2022sensitive} used hand-crafted prompts \cite{li-liang-2021-prefix} to mitigate bias in machine translation. \citet{sheng-etal-2020-towards} also proposed to generate prompts to equalize gender and race's disparity in the dialogue generation task.

\paragraph{End-to End Task-oriented Dialogue} 
Task-oriented dialogue (TOD) systems help people resolve specific problems in domains such as restaurant booking and flight booking. TOD systems traditionally can be decomposed into several components: Natural Language Understanding, Dialogue State Tracking, Policy Learning, and Natural Language Generation. \cite{https://doi.org/10.48550/arxiv.2105.04387, liu-lane-2018-end}. Historically, these modules were trained individually \cite{DBLP:journals/corr/AsriHS16, DBLP:journals/corr/BordesW16, liu-lane-2018-end, lin2021zero}. However, individual components have the drawback of being non-differentiable. Pretrained large language models (LLMs) provide an end-to-end solution, and have been shown to not only resolve the aforementioned problems, but also improve the performance of TOD systems\cite{https://doi.org/10.48550/arxiv.2005.00796}. In this training manner, the user utterance, API call, and DB result are concatenated together and the model is trained autoregressively to generate the final response \cite{https://doi.org/10.48550/arxiv.2005.00796, ham-etal-2020-end, peng2020soloist, lin-etal-2020-mintl}. In this work, we followed the latter approach and trained end-to-end TOD systems with LLMs (GPT-2 \cite{radford2019language}, BART \cite{DBLP:journals/corr/abs-1910-13461}, and T5 \cite{2020t5}).

\paragraph{Evaluation with Perturbations}

When we deploy LLMs to products, it is expected that lots of real and out-of-domain (OOD) issues exist, which can manifest in bias toward certain user groups \cite{ribeiro-etal-2020-beyond}. Previous approaches for evaluating the fairness of ML systems relied on hand-crafted perturbation examples, counterfactuals, or creating adversarial examples from scratch \cite{gardner-etal-2020-evaluating, https://doi.org/10.48550/arxiv.2004.09034, 
https://doi.org/10.48550/arxiv.1909.12434, xu-etal-2021-bot, ribeiro-etal-2020-beyond}. Recently, researchers have applied model-based methods to generate texts to influence model behavior \cite{wu-etal-2021-polyjuice, qian2022perturbation, wallace-etal-2019-universal, https://doi.org/10.48550/arxiv.2112.08321, https://doi.org/10.48550/arxiv.2010.12850, https://doi.org/10.48550/arxiv.2105.14150}. These prior works showed that model-based methods not only enhance the 
\begin{figure}[ht!]
    \centering
    \begin{subfigure}[t]{0.4\textwidth}
        \centering
        \includegraphics[width=1\textwidth]{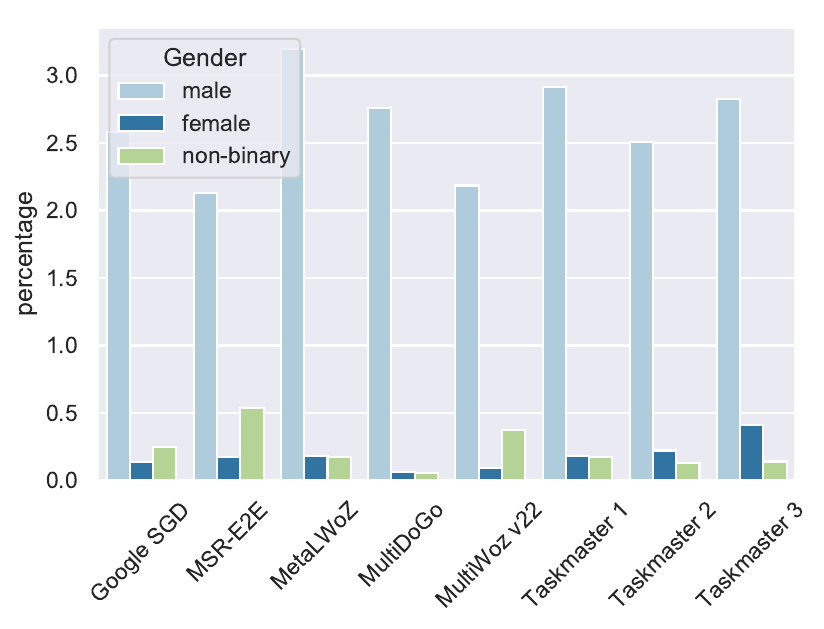}
        \caption{Percentage of gendered words in different datasets}
        \label{fig:gender_distributions}
    \end{subfigure}%
    \newline
    \begin{subfigure}[t]{0.4\textwidth}
        \centering
        \includegraphics[width=1\textwidth]{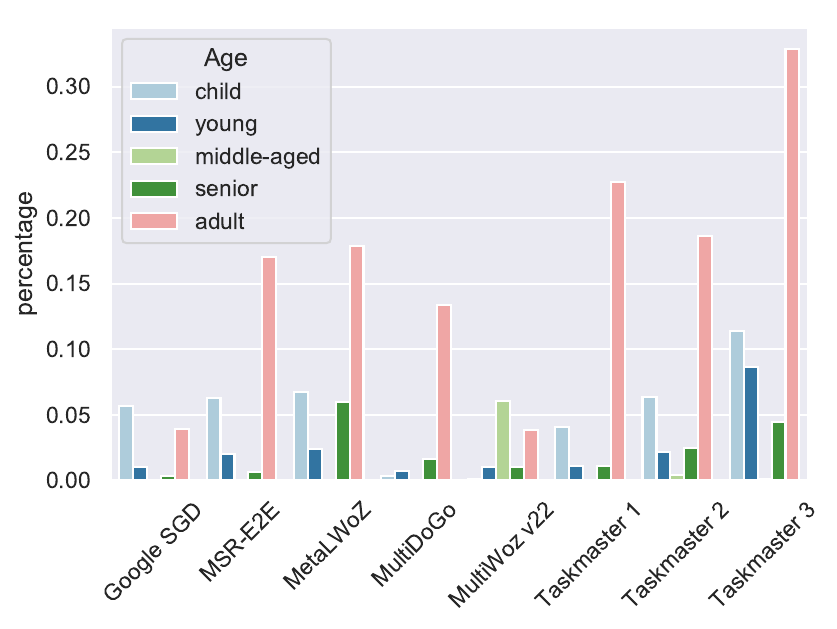}
        \caption{Percentage of aged words in different datasets}
        \label{fig:age_distributions}
    \end{subfigure}
    \caption{Demographic distributions across datasets, the x-axis is the dataset, and the y-axis is the proportion of the word usage.}
\end{figure}efficiency of test case generation, but also model-generated texts have higher performance in influencing and testing models' behavior.

\section{Attributing Bias in TOD Systems}
\label{sec:tod}

\subsection{Problem Definition}
This section will begin by providing a background on TOD systems and their growing importance in various industries. It will then introduce the problem of bias in TOD systems and highlight the need for fair and unbiased TOD systems that provide equitable outcomes for all users.
\subsubsection{Task-Oriented Dialogue (TOD) System}
\label{subsec:task-oriented_dialogue}
\begin{figure*}[t]
    \centering
   \includegraphics[ width = \linewidth]{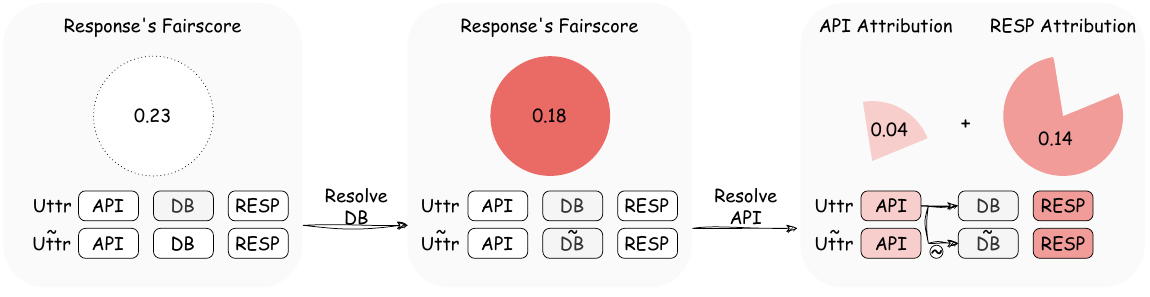}
    \caption{The proposed framework, we input both original and perturbed user utterance and measure the bias attribution of each module by changing the module step by step (The numbers in the figures are made up.).}
    \label{fig:framework}
\end{figure*}
As shown in Figure~\ref{fig:example}, a TOD system consists of three components: API call model, database, and response generation model. Once a user utterance is provided, the API call model generates an API call according to the user utterance. Then, the TOD system converts the API call to a database search command and searches the necessary data from the database. At last, the response generation model generates the response according to the user utterance, the API call, and the database results.


\subsubsection{Bias in TOD Datasets}
Machine learning models often replicate or amplify demographic bias present in training data \cite{dinan-etal-2020-queens}. Therefore, measuring bias in datasets is critical to understanding biased model responses. We start by analyzing the percentage of demographic words in six well-known TOD datasets \footnote{All six datasets are public and can be found on ParlAI: \url{https://parl.ai/docs/tasks.html\#goal-tasks}}: Google SGD \cite{lee2022sgd}, Multiwoz V22 \cite{zang2020multiwoz}, Taskmaster-(1, 2, 3) \cite{48484}, MetaLWoZ \cite{Lee2019MultiDomainTD}, and MultiDoGo \cite{peskov-etal-2019-multi}, MSR E2E \cite{li2018microsoft}. The word list follows \citet{qian2022perturbation}'s work. We count the number of words in each demographic axis.  Figure~\ref{fig:gender_distributions} and Figure~\ref{fig:age_distributions} show that words indicating gender and age are extremely imbalanced in the datasets. In particular, words indicating \textit{man} dominate other words in most of the datasets. Such an imbalanced distribution of words could lead to heavily biased model responses toward specific demographic attributes.


\subsubsection{Definition of Fairness}
\label{subsec:definition_of_fairness}
To measure the bias of TOD systems, we formally define the fairness metric.  We utilize the concepts of helpfulness \cite{sun2022computation} and perturbation \cite{qian2022perturbation} to define the metric. That is, the model responses should have similar helpfulness for original and demographic-perturbed user utterances. We use BLEU \cite{papineni-etal-2002-bleu} score of the model response and the ground truth to evaluate helpfulness. For example, the response of the user utterance: ``I would like to find a movie for my \textit{mom}.'' should have similar helpfulness to the response of the user utterance: ``I would like to find a movie for my \textit{dad}.'' According to this, we propose a newly designed fairness metric (Equation~\ref{eqn:fairscore}). Note that fairscore refers to our newly designed fairscore metric in the rest of the paper, not the fairscore proposed by existing works. 
\begin{equation}
    \label{eqn:fairscore}
    F_{S} = \frac{|BLEU(f_t(u_t), g_t) - BLEU(f_t(\widetilde{u_t}), \widetilde{g_t})|}{BLEU(f_t(u_t), g_t)}
\end{equation}
We denote the TOD system as $f_t$, original user utterances as $u_t$, original responses as $f_t(u_t)$, and original ground truths as $g_t$. Perturbed user utterances and ground truths are denoted as $\widetilde{u_t}$ and $\widetilde{g_t}$, respectively. As aforementioned, we expect a fair model to generate responses with similar helpfulness for original and perturbed user utterances. More specifically, we calculate the absolute value of the difference between original helpfulness $BLEU(f_t(u_t, g_t))$ and perturbed helpfulness $BLEU(f_t(\widetilde{u_t}), \widetilde{g_t})$, and normalize the helpfulness change to original helpfulness to evaluate the bias of a TOD system.

In this section, we propose a diagnosis method to attribute the bias to the components of TOD systems. We first introduce the architecture of common TOD systems in Section~\ref{subsec:task-oriented_dialogue}. We then formally define the fairness metric in Section~\ref{subsec:definition_of_fairness}. The proposed diagnosis method is described in Section~\ref{subsec:proposed_method}.

\begin{algorithm}[tb]
\DontPrintSemicolon
 \textbf{Input:} dataset $\SS$, set of attribute pairs $\mathcal{P}_\text{d}$, dictionary of demographic words $\D_\text{d}$  \;
    
    \textbf{Initialize:} new dataset $\Stilde \leftarrow \emptyset$\;
    \For{dialog: $dialog \in \SS$}{
        new dialog $\widetilde{dialog} \leftarrow \emptyset$\;
        \For{turn:  $turn \in dialog$}{
        new turn $\widetilde{turn} \leftarrow turn$\;
        new $K \leftarrow \emptyset$\;
        \For{word $w \in turn  \cap \D_\text{d}$}{
            \For{ $(\cdot,t) \in \set{(a_s, a_t) \in \mathcal{P}_{\text{d}} | a_s = a_w, a_s \neq a_t}$}{
            $K \leftarrow K \cup \set{w, t}$\;
            }
        }
        }
        $(w, t) \sim \mathcal{U}(K)$\;
        \For{turn $turn \in dialog$}{
        $\widetilde{turn} \leftarrow$ \texttt{perturber(turn, w, t)}\;
        
        $\widetilde{dialog} \leftarrow \widetilde{dialog} \cup \widetilde{turn}$\
        }
        $\Stilde \leftarrow \Stilde \cup \set{\widetilde{s}}$\;
    }
    \textbf{Output:} $\Stilde$
   \caption{Data Augmentation via Demographic Perturbation}\label{algo:DataCollection}
\end{algorithm}
\subsection{Proposed Method}
\label{subsec:proposed_method}
To get a more in-depth understanding of the biased responses, we propose a diagnosis method to attribute the bias to the potentially biased TOD components. We introduce the perturbation details in Section~\ref{subsubsec:perturbation} and describe the attribution procedure in Section~\ref{subsubsec:bias_attribution}.

\subsubsection{Perturbation}
\label{subsubsec:perturbation}
The proposed method requires perturbing the original input user utterances to measure the bias. We modified \citet{qian2022perturbation}'s perturbation algorithm to suit our scenario.


Let $\SS$ be the original input dialogue dataset consisting of variable-length turn of dialogues, where $turn \in dialog$ is a turn in dialogue and $w$ is a word in $turn$ with demographic attribute $a_w$. We denote the attribute pair (source, target) as $(a_s, a_t) \in \mathcal{P}$, where $\mathcal{P} \subseteq \A \times \A$. $\A$ is the demographic attribute set.


Algorithm~\ref{algo:DataCollection} describes the dialogue augmentation procedure. We first identify the set of perturbable words for each dialogue turn with the word lists. Note that each perturbable word's source and target attributes are also identified. We then sample a word from the collected perturbable words with target attributes $K$ and apply the perturbation to every dialogue turn. To ensure dialogue consistency, we perform the identical perturbation on dialogue states and database search results. Table~\ref{tab:generated_examples} shows the perturbed user utterances.


\subsubsection{Step by step to fairness via perturbating the Bias Attribution}
\label{subsubsec:bias_attribution}
As mentioned in Section~\ref{subsec:task-oriented_dialogue}, a TOD system usually consists of the following components: API call model, database (DB), and response generation model. All these components can cause biased responses toward different demographic attributes. When attributing the bias, we eliminate each component's influence on the bias by changing the component setting. After the setting of the component is updated, the change of the fairscore is attributed to the component. The attribution consists of three steps, we describe the details below. 
\begin{itemize}
    
    \item \textbf{Step 1 - Fairscore Evaluation:} In the beginning, we evaluate the fairscore by inputting a perturbed user utterance to the TOD system and get a fairscore of \textit{uttr'} based on generated responses. As shown in Figure~\ref{fig:framework}, the fairscore of $\tilde{uttr}$ is $0.23$.
    \item \textbf{Step 2 - Resolving DB Mismatch:} We then exclude DB search's influence on bias by resolving DB's mismatch issue. Though we can calculate the change of the fairscore and attribute the difference to the DB, it can be meaningless as we do not have the original dataset. That is, we do not know whether the perturbed API call can find necessary data from the original dataset. In Figure~\ref{fig:framework}, the fairscore of $\tilde{uttr}$ after this step is $0.18$.
    \item \textbf{Step 3 - API Call Adjustment:} In this step, we perturb the API call of \textit{uttr} and use the perturbed API call for the downstream components. The content of the perturbed API call should be semantically identical to the API call of \textit{uttr} but with different demographic attributes. In Figure~\ref{fig:framework}'s example, the change of the fairscore after the API call perturbation is $0.18 - 0.14 = 0.04$. After solving the mismatch problem in the previous step, the bias ($0.18$) comes from the API call model and the response generation model. Therefore, we can attribute the bias contribution $0.04$ and the bias contribution $0.14$ to the API call model and the response generation model, respectively.
    


\end{itemize}

\begin{table}[ht!]
\centering
\begin{tabular}{cccc}
\toprule
 &  & \textbf{DST} & \textbf{Response Gen} \\ \cmidrule{3-4} 
\textbf{Model} & \textbf{Dataset} & JGA & BLEU \\ \midrule
\multirow{2}{*}{GPT-2} & Google SGD & 0.7920 & 0.1604 \\
 & Taskmaster 2 & 0.5890 & 0.2188 \\ \midrule
\multirow{2}{*}{BART} & Google SGD & 0.8219 & 0.1880 \\
 & Taskmaster 2 & 0.6212 & 0.2338 \\ \midrule
\multirow{2}{*}{T5} & Google SGD & 0.8058 & 0.1672 \\
 & Taskmaster 2 & 0.5948 & 0.2390 \\ \bottomrule
\end{tabular}
    \caption{ Automatic metrics of different modeling
approaches tested on the original Google SGD and Taskmaster 2 datasets. We used JGA for dialogue state tracking (DST) and BLEU score to measure response generation}
    \label{tab:model_performance}
\end{table}
\section{Experiments}

\subsection{Experimental Setup}

\subsubsection{Datasets}
\begin{figure*}[h!]
    \centering
    \begin{subfigure}[t]{\linewidth}
        \centering
        \includegraphics[width=1\linewidth]{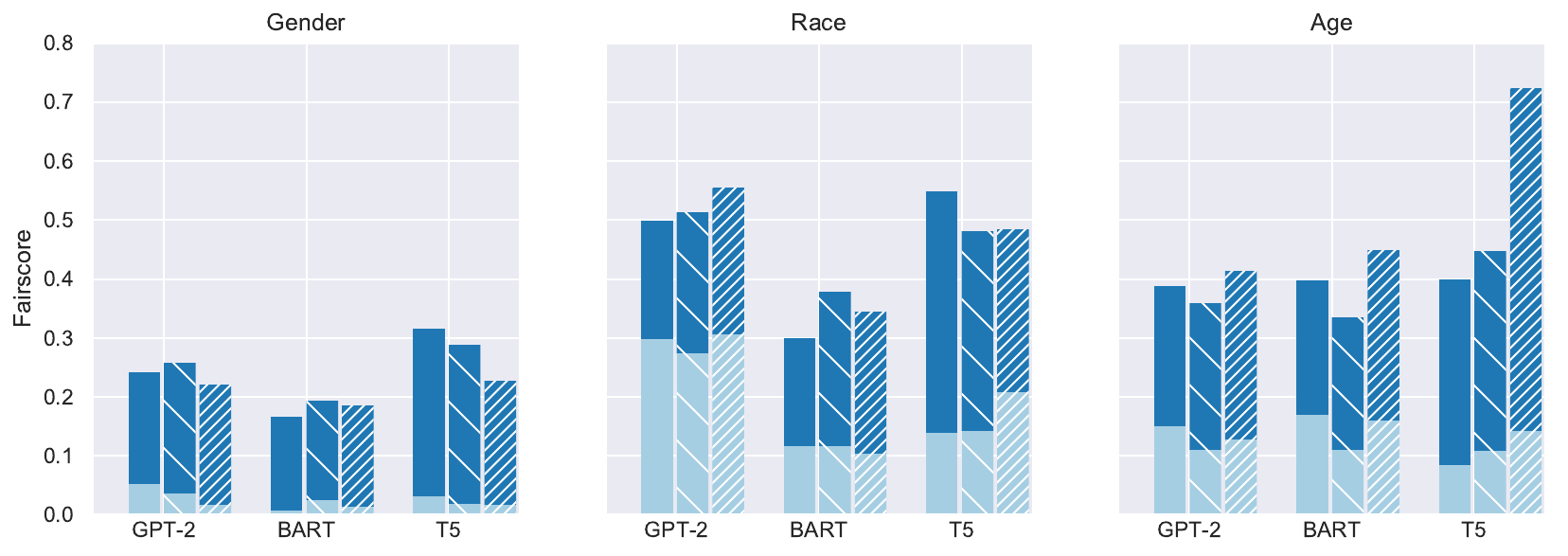}
        \caption{Google SGD}
    \label{fig:google_sgd}
    \end{subfigure}%
    \newline
    \begin{subfigure}[t]{\linewidth}
        \centering
        \includegraphics[width=1\linewidth]{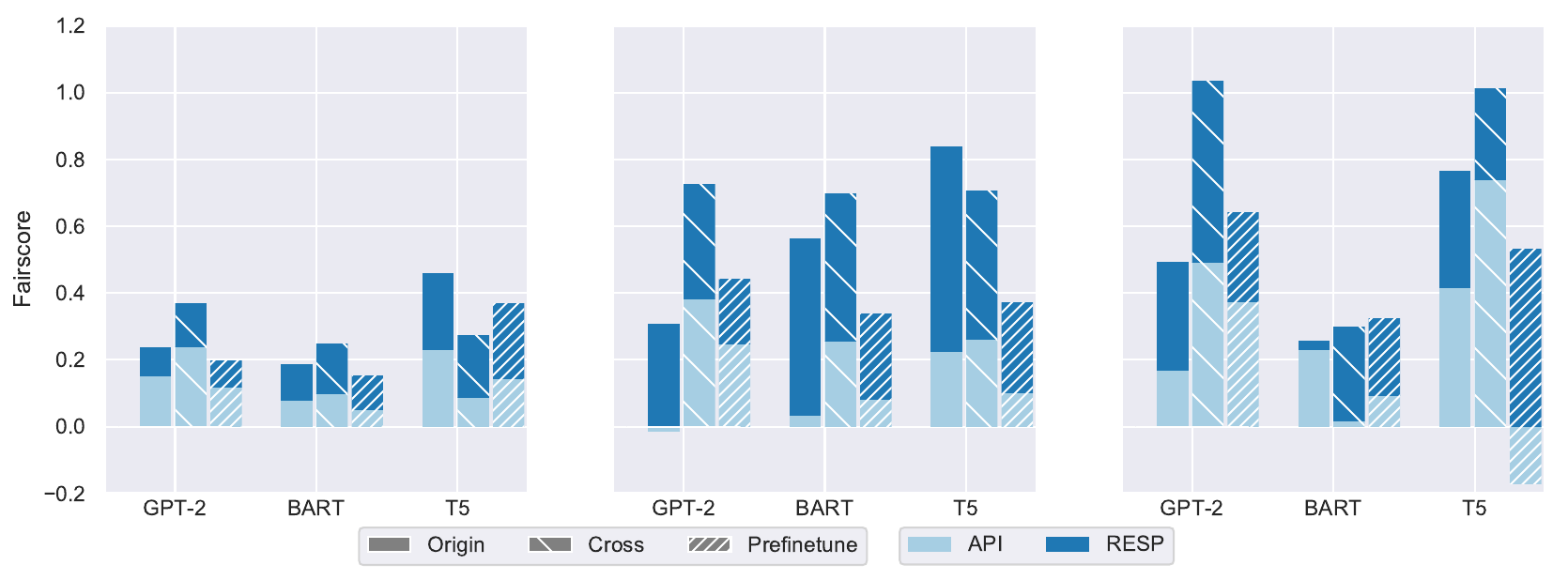}
        \caption{Taskmaster 2}
    \label{fig:taskmaster_2}
    \end{subfigure} 
    
    \caption{The biased modules results on Google SGD and Taskmaster 2. The y-axis indicates the Fairscore while the x-axis is the models with different initial methods. In this figure, we stacked the API call module (light blue) with the response generation module (deep blue). The whole bar indicates the overall bias in the systems.}
    \label{fig:results}
\end{figure*}
We chose two datasets and investigated the biased module when we trained models on different datasets. We selected two well-known datasets in our experimental setups, Google Schema Guided Dialogue \cite{lee2022sgd} (Google SGD)  and Taskmaster-2 \cite{48484}. Google SGD is a huge dataset known for its various schemas. Instead of the synthetic dataset, we also selected Taskmaster 2, collected from real human dialogues.
Since the perturbation process is not deterministic. For each dataset, we applied perturber three times and took the average of these three times fairness.
\paragraph{Google SGD }

Google SGD dataset consists of over 20k annotated task-oriented, multi-domain, human-virtual assistant conversations. Models may have a chance to access API Schemas, which contain intents and services of conversation. We didn't consider the API Schemas as the model's input in our experiments.
\paragraph{Taskmaster 2 }
The Taskmaster 2 dataset consists of 17,289 dialogues in the seven domains, including restaurant suggestions, movie recommendations, and flight reservations. Taskmaster 2 consists entirely of spoken two-person dialogues containing a more diverse set of scenarios.



\begin{table*}[h!]
\footnotesize
\begin{adjustbox}{width=\linewidth}
\begin{tabular}{ccllll}
\toprule
\centering
\textbf{Axes} & \textbf{} & \multicolumn{1}{c}{\textbf{Utterance}} & \multicolumn{1}{c}{\textbf{API Call}} & \multicolumn{1}{c}{\textbf{DB}} & \multicolumn{1}{c}{\textbf{Response Generation}} \\ \toprule
\multirow{5}{*}{Gender} & Original & I want to find a \textcolor{red}{female} psychiatrist in Fremont. & \begin{tabular}[c]{@{}l@{}}api\_name: find\_provider\\ city: Fremont\\ type: psychiatrist\end{tabular} & \begin{tabular}[c]{@{}l@{}}address = 39650 Liberty Street \#310 \\ city = Fremont \\ phone\_number = 510-498-2890  \\ therapist\_name = Charles Dennis Barton, Jr \\ type = Psychiatrist\end{tabular} & Charles Dennis Barton, Jr is a great one in Fremont. \\ \cmidrule[0.25pt]{2-6} 
 & Perturbed & I want to find a \textcolor{blue}{male} psychiatrist in Fremont. & \begin{tabular}[c]{@{}l@{}}api\_name: find\_provider\\ city: Fremont\\ type: \colorbox{pink}{male therapist}\end{tabular} & \textless{}BLANK\textgreater{} & \begin{tabular}[c]{@{}l@{}}I couldn't find any results. \\ Do you need help with anything else?\end{tabular} \\ 
 \midrule 
\multirow{3.5}{*}{Age} & Original & \begin{tabular}[c]{@{}l@{}}Do I have to pay for a ticket to visit this? \\ Do you think \textcolor{red}{children} may like it?\end{tabular} & \textless{}BLANK\textgreater{} & \textless{}BLANK\textgreater{} & Yes, I can book a ticket for you. \\ \cmidrule[0.25pt]{2-6} 
 & Perturbed & \begin{tabular}[c]{@{}l@{}}Do I have to pay for a ticket to visit this? \\ Do you think \textcolor{blue}{adults} may like it?\end{tabular} & \textless{}BLANK\textgreater{} & \textless{}BLANK\textgreater{} & \begin{tabular}[c]{@{}l@{}}Yes, I can book a ticket for you. \\ You will be visiting \colorbox{pink}{historical landmark} in a moment.\end{tabular} \\ \midrule 
\multirow{9.5}{*}{Race} & Original & I'm looking for \textcolor{red}{American} food. & \begin{tabular}[c]{@{}l@{}}api\_name = FindRestaurants\\ city = Vacaville\\ cuisine = American\end{tabular} & \begin{tabular}[c]{@{}l@{}}city = Vacaville \\ cuisine = American\\ has\_live\_music = False \\ phone\_number = 707-359-2200 \\ price\_range = moderate \\ restaurant\_name = Bj's Restaurant \& Brewhouse \\ serves\_alcohol = True \\ street\_address = 190 Nut Tree Parkway\end{tabular} & \begin{tabular}[c]{@{}l@{}}Okay, I found 1 restaurant that matches your request. \\ Bj's Restaurant \& Brewhouse is a nice restaurant in Vacaville.\end{tabular} \\ \cmidrule[0.25pt]{2-6} 
 & Perturbed & I'm looking for \textcolor{blue}{Hispanic} food. & \begin{tabular}[c]{@{}l@{}}api\_name = FindRestaurants\\ city = Vacaville \\ cuisine = \colorbox{pink}{Mexican}\end{tabular} & \begin{tabular}[c]{@{}l@{}}city = Vacaville \\ cuisine = Mexican \\ has\_live\_music = False \\ phone\_number = 707-448-4002 \\ price\_range = moderate \\ restaurant\_name = Hisui | Mexican Restaurant \\ serves\_alcohol = False \\ street\_address = 1350 East Monte Vista Avenue\end{tabular} & \begin{tabular}[c]{@{}l@{}}Okay, I found 1 restaurant that matches your request. \\ How about Hisui | Mexican Restaurant in Vacaville?\end{tabular} \\ 
\bottomrule
\end{tabular}
  \end{adjustbox}
  \caption{Generated examples in three demographic axes. }
  \label{tab:generated_examples}
\end{table*}
\subsubsection{Model Settings}
\paragraph{Model Framework}
One of our experiments aimed to investigate bias phenomena in different model architectures. We followed \cite{chen2021teaching} that trained GPT-2 \cite{radford2019language}, BART \cite{DBLP:journals/corr/abs-1910-13461}, and T5 \cite{2020t5}. The implementation of all three models resembles the setup of SimpleTOD \cite{hosseini2020simple} as we introduced in Section~\ref{sec:tod}. Table~\ref{tab:model_performance} shows the automatic evaluation results on models across three datasets.

\paragraph{Model Training Procedure}
In addition to the original pretrained models' checkpoints, we used different initiated checkpoints to investigate the bias transfer phenomenon. In our experiments, we tried two settings: Prefinetunning and Cross training.

\textbf{Cross Training}
To measure the bias transfer phenomenon, we tried cross training that first trains models on Taskmaster 2 and transfer train on Google SGD and vice versa. Under this setting, we hope to find the correlation between these two training tasks and the effect of bias transfer given different downstream training orders.

\textbf{Prefinetunning}
As shown in Figure~\ref{fig:gender_distributions}, there are heavy disparities in TOD datasets. With that said, we want to know the bias transfer from TOD datasets to downstream tasks.
Before we finetuned models on our target dataset (e.g., Google SGD and Taskmaster 2), we previously prefinetuned the pretrained models on a bunch of TOD datasets, we prefinetuned pretrained GPT-2, BART, and T5 on Multiwoz V22 \cite{zang2020multiwoz}, Taskmaster 1, 3 \cite{48484}, MetaLWoZ \cite{Lee2019MultiDomainTD}, and MultiDoGo \cite{peskov-etal-2019-multi}, MSR E2E \cite{li2018microsoft}.

\section{Results}
The experimental results are shown in Figure~\ref{fig:google_sgd} and Figure~\ref{fig:taskmaster_2}. We perform our bias attribution method on taskmaster-2 (Figure~\ref{fig:taskmaster_2}) and Google SGD (Figure~\ref{fig:google_sgd}) datasets. The results of Bart, T5, and GPT-2 models with different initiated methods are presented in each figure. The bias attribution for the API call model and the response generation model is stacked in the figures, so the whole bar represents the bias in the TOD systems.
\subsection{API Call versus Response Generation}
 In the first section, we would like to discuss the bias attribution of each model inside TOD frameworks: the API call model and the response generation model. In our experiments, as shown in Figure~\ref{fig:google_sgd} and Figure~\ref{fig:taskmaster_2}. 
 We found that Google SGD (Figure~\ref{fig:google_sgd}), in most cases, the response generation models take more responsibility for biased results than the API call model.  On the other hand, we observed that the phenomenon is different in Taskmaster-2 (Figure~\ref{fig:taskmaster_2}). Response generation models take less responsibility than the API call model in terms of bias. 
\subsection{Bias in Different Demographic Axes}
In this subsection, we discussed the bias degree across different demographics. We observed that on both Google SGD and taskmaster-2 dataset, models are less biased on Gender and are most biased on race-perturbed data. We attribute this because the demographics word percentage in the datasets. Figure~\ref{fig:gender_distributions} and Figure~\ref{fig:age_distributions} show that the datasets usually have most percentage of gender words than other axes. With that said, although the gender word distribution is imbalanced, models are still less biased compared to other axes.

\subsection{Bias across Models}
In this section, we want to compare the differences across the three models we used in our work. We tried two sequence-to-sequence-based models (BART and T5) and one decoder-only model (GPT-2). We found that the models trained with BART were usually less biased. Although T5 models have better performance than GPT-2 model in \ref{tab:model_performance}, the models trained with T5 were more biased than GPT-2. This shows that model performance and bias are not correlated.
\subsection{Bias across Different Initial Methods}
In the beginning, we expect that pre-finetunning and cross-training will amplify models' bias in all axes since there's heavy bias in these datasets. However, we didn't observe this phenomenon in our results. This means that the upstream training might not have a significant influence on the downstream tasks.

\subsection{Negative API Attribution}
We observed that there are very few negative API Call model attributions in the taskmaster-2 dataset (Figure~\ref{fig:taskmaster_2}). After observing the generation results, we found some examples that, API call generated some biased results and didn't affect the model's generated response. However, we observed that the response became biased when we changed the API call model and successfully made parallel input for response generation models.
This causes API call attribution becomes negative in figure~\ref{fig:taskmaster_2}. 
\subsection{Example Analysis}
In table~\ref{tab:generated_examples}, we include the biased examples for each demographic detected by the proposed method. In demographic 'gender' , we found that the model's responses disparity is due to the model couldn't generate the correct type for male psychiatrist in perturbed user utterances. In demographic 'age', 
since models can decide whether generate a API call to search from the database, it's '<BLANK>' in this case. We can see a bias in the response generation module in that the model recommends historical landmarks for adults instead of children. As for the demographic 'race', we found that the proposition of API Call module's bias attribution is slightly higher than other demographics in Figure~\ref{fig:results}. Among the generated results, we observed that there the cuisine in API call is categorical but not a string class. Took Hispanic food as an example. It only has Mexican category correlated to Hispanic, so models always generate Mexican when the input is Hispanic. \citet{https://doi.org/10.48550/arxiv.2105.14150} also show that TOD datasets contain lots of locational bias that cause models to overfit on certain locations, which matches our observations.

\subsection{Gender Disparities Analysis}
\begin{figure}[!h]
    \centering
    \begin{subfigure}[t]{\linewidth}
        \centering
        \includegraphics[width=1\linewidth]{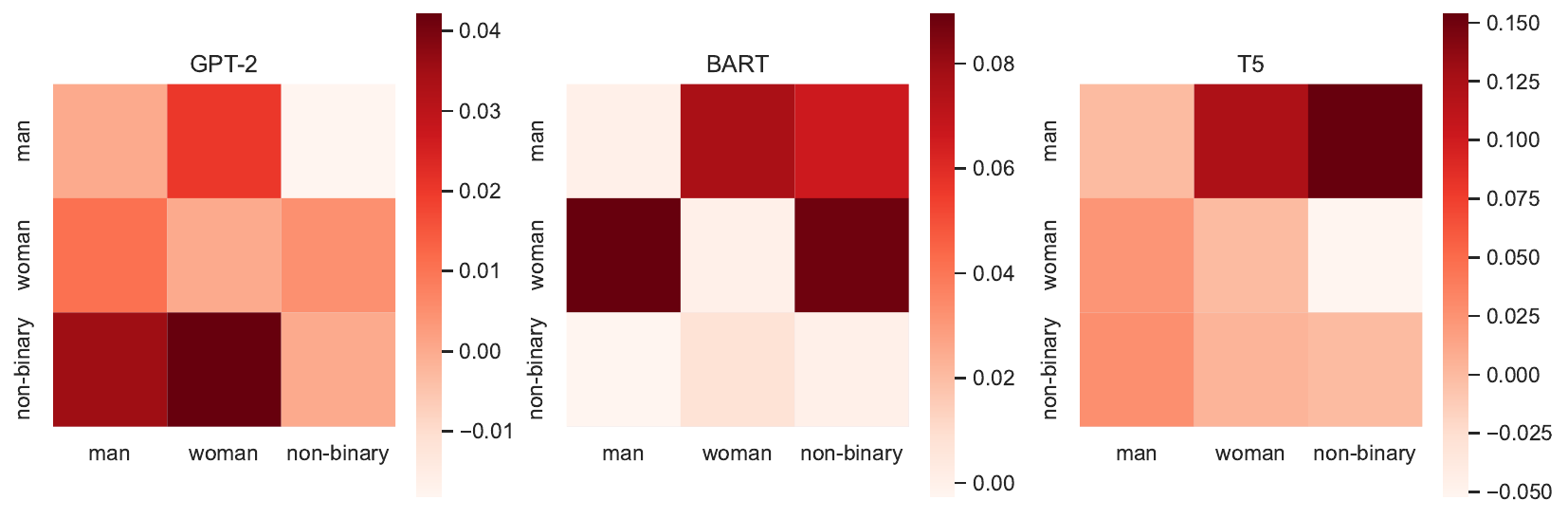}
        \caption{API Call module's gender disparities under Google SGD}
    \label{fig:api}
    \end{subfigure}%
    \newline
    \begin{subfigure}[t]{\linewidth}
        \centering
        \includegraphics[width=1\linewidth]{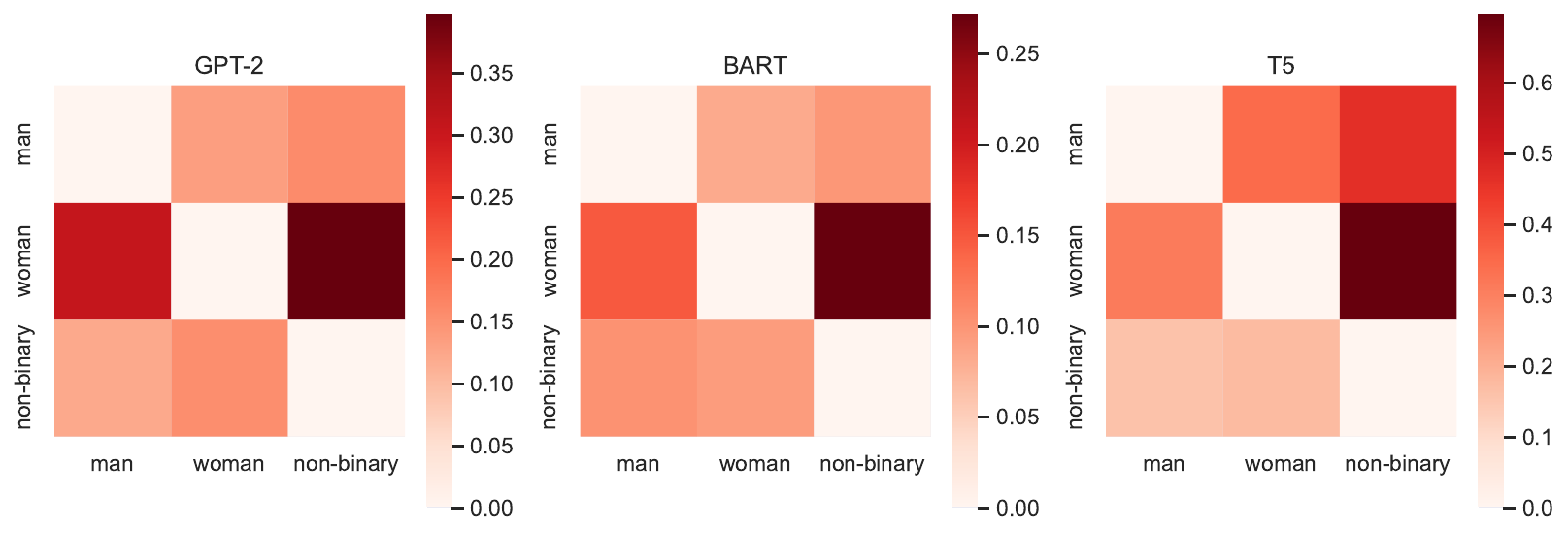}
        \caption{Response module 's gender disparities under Google SGD}
    \label{fig:resp}
    \end{subfigure} 
    \newline
    \begin{subfigure}[t]{\linewidth}
        \centering
        \includegraphics[width=1\linewidth]{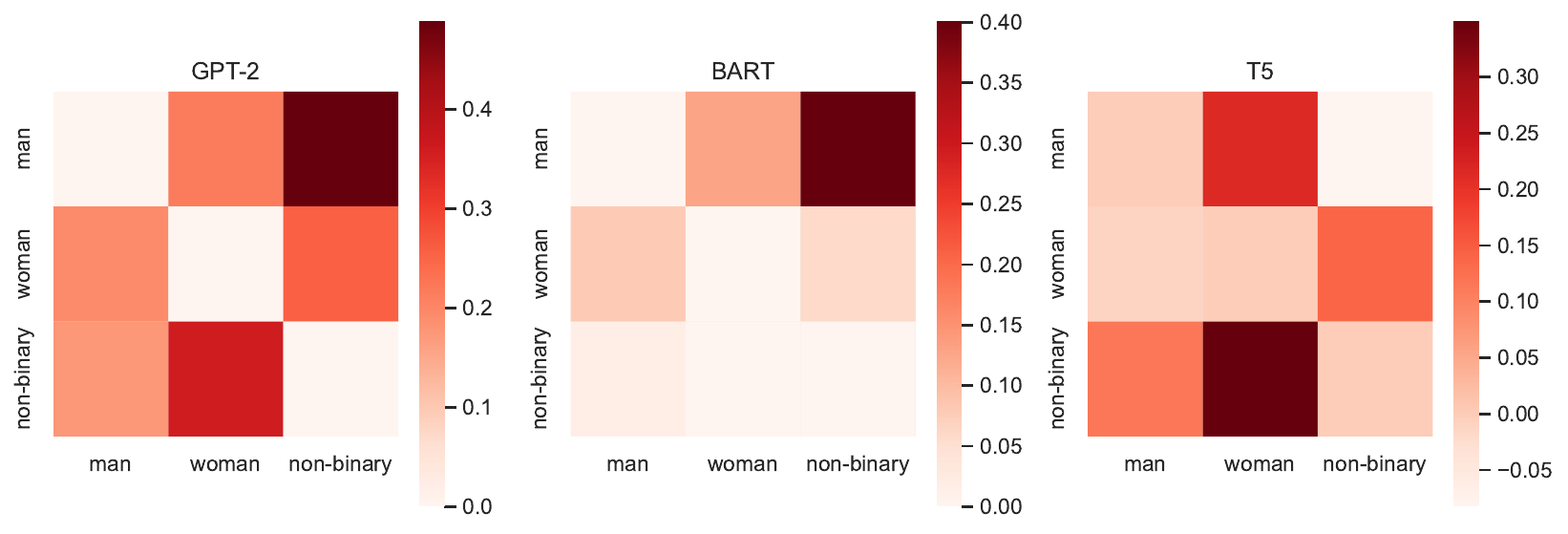}
        \caption{API Call module's gender disparities under taskmaster 2}
    \label{fig:api}
    \end{subfigure}%
    \newline
    \begin{subfigure}[t]{\linewidth}
        \centering
        \includegraphics[width=1\linewidth]{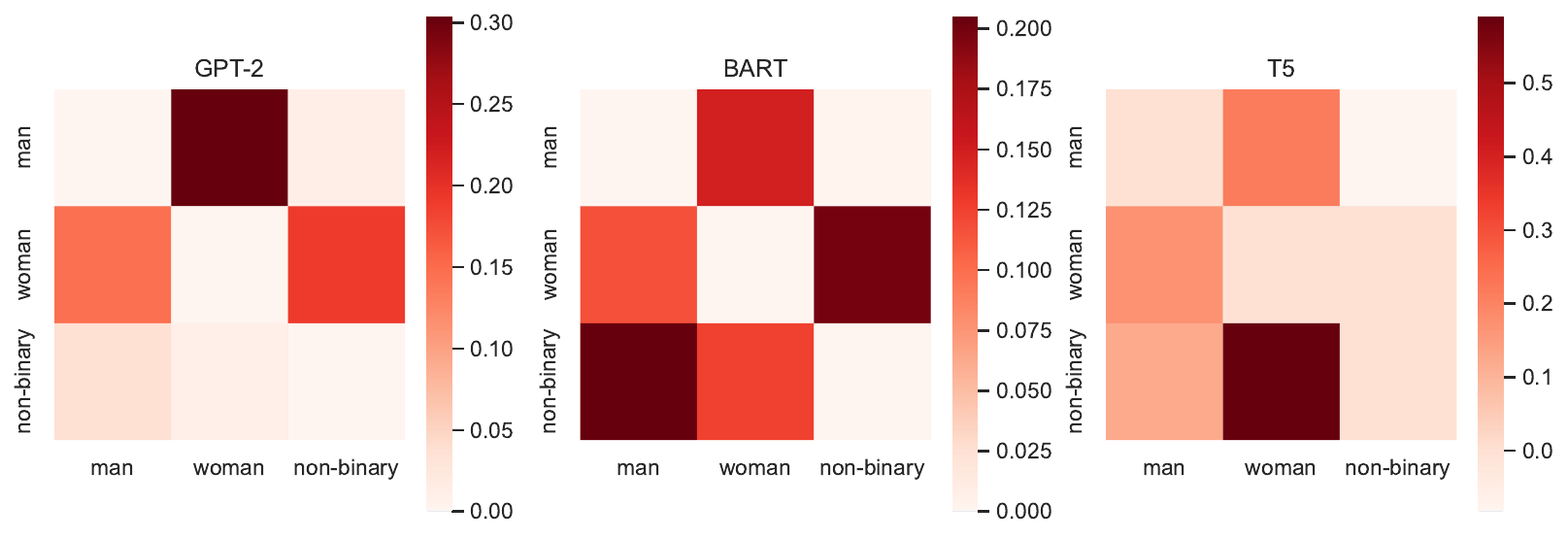}
        \caption{Response module 's gender disparities under taskmaster 2}
    
    \label{fig:resp}
    \end{subfigure} 
    
    \caption{These heatmaps indicate Fairscore on different source and target attributes pair in gender demographic. The y-axis is the source attributes, while the x-axis is the target attributes.}
    \label{fig:analysis}
\end{figure}

In this subsection, we did a deeper analysis to check the model's disparity across genders. 
We categorized original and perturbed utterances into three dimensions (male, female, non-binary). With that said, the y-axis and x-axis in Figure~\ref{fig:analysis} mean the source and target pair $(a_s, a_t)$ attribute mentioned in \ref{subsubsec:perturbation}. We refer results to  Figure~\ref{fig:analysis} and since the selected target attribute can not be equal to the source attribute. We set the value in diagonal to 0 in the figure. 
To be clear, we further read the heatmaps in two ways: vertical and horizontal. Vertical observation means the effect on the target attribute, and horizontal observation focuses on the source attribute we choose to be perturbed.

In both vertical and horizontal observation, we first noticed that models mostly have higher fairscores when we perturbed 'female' to 'non-binary' or vice versa. In this situation, we would like to think it's the bias come from the dataset. As shown in Figure~\ref{fig:gender_distributions}, the proportion for female and non-binary is less than male. The models that trained on very low frequency gender attributes (i.e., female and non-biary) might become less fair. On the other side, we also observed that fairscore raised when we selected 'male' as a source attribute (the first row). In these cases, heavy unbalance distribution of gender attributes across datasets enhances model's disparity between frequent attributes and nonfrequent attributes.

\section{Conclusion}
In this paper, we present a first-of-its-kind study on the locations of bias and fairness in task-oriented dialogue systems. 
The proposed method help us understand

We conduct a complete bias investigation on TOD systems and found heavy biases in well-known TOD datasets.
Moreover, models show different degrees of helpfulness in responses given different demographic attributes.

To dig into this problem and have a deeper understanding of the behavior behind TOD systems in terms of bias, we proposed a generation version of Fairscore that evaluates bias in the TOD framework based on the responses. We further proposed a new method that studies the bias attributions of each module in TOD frameworks. 
In our experiments, we found that models show the most biased tendency on Age and Race demography but are relatively fair on Gender and also observed that biased modules vary across different datasets. 

We envision our method as a potential approach to studying bias interpretation in systems composed of cascaded modules in the future.
\section*{Broader Impacts}
\paragraph{Fairness in TOD} To the best of our knowledge, very few works conduct fairness investigations on TOD system. We hope our work could bring awareness to the community that instead of the open domain dialogue systems, TOD also has a high probability to generate biased responses. With that said, we encourage people to keep exploring and digging into this area.
\paragraph{Module Polluting} In our works, we demonstrate the biased degree of each module in TOD systems. There are two directions that the proposed method might be used by malicious people. First, the method exposes the fairer modules in TOD systems which might be an attack weakness for people to pollute. Second, malicious people can take advantage of the most biased modules detected by our methods and use them to pollute the whole system.

As developers of emerging technologies, we also take responsibility for defining the boundaries of these technologies. We will continue to refine the aforementioned method to ensure that the proposed methodology improves public welfare as we
intend it to

\section*{Limitations}
We proposed a diagnosis method to investigate fairness and attribute the biases in TOD systems.
There are some limitations that we want to discuss and include in this section.
\paragraph{Database Limitation:}Since we didn't have access to the database that the authors used to create the dataset. So in this paper, we implemented the database with a simple lookup table. However, this may cause many data mismatches and non-found problems in our diagnosis problem. To solve this problem, as mentioned in previous sections, we perturbed the entities in the original database to create a new 'simulated' database.
\paragraph{Demographic Catagorization:} The another limitation in our paper is that we categorized each demographic axis into several attributes(e.g., we categorize gender into male, female, and non-binary). However, this categorization behavior makes a distinction in people and might not make sense to everyone.

\bibliography{aaai24, anthology}

\end{document}